\documentclass[
]{ceurart}

\usepackage{listings}
\usepackage{natbib}
\usepackage{graphicx}
\usepackage{caption}
\usepackage{subcaption}
\usepackage{array}
\usepackage{colortbl}

\begin{document}

\copyrightyear{2024}
\copyrightclause{Copyright for this paper by its authors.
  Use permitted under Creative Commons License Attribution 4.0
  International (CC BY 4.0).}

\conference{The 2nd Workshop \& Challenge on Micro-gesture Analysis for Hidden Emotion Understanding, Aug 3--9, 2024, Jeju, South Korea}

\title{Micro-gesture Online Recognition using Learnable Query Points}


\author[1]{Pengyu Liu}[%
orcid=0000-0002-3396-3108,
email=lpynow@gmail.com,
]

\author[1]{Fei Wang}[%
orcid=0009-0004-1142-6434,
email=jiafei127@gmail.com,
]

\author[4]{Kun Li}[%
orcid=0000-0001-5083-2145,
email=kunli.hfut@gmail.com,
]
\cormark[1]

\author[1]{Guoliang Chen}[%
orcid=0009-0002-7984-3184,
email=guoliangchen.hfut@gmail.com,
]

\author[1]{Yanyan Wei}[%
orcid=0000-0001-8818-6740,
email=weiyy@hfut.edu.cn,
]

\author[1]{Shengeng Tang}[%
orcid=0000-0001-6313-2543,
email=tangsg@hfut.edu.cn,
]

\author[4]{Zhiliang Wu}[%
orcid=0000-0002-6597-8048,
email=wu_zhiliang@zju.edu.cn,
]

\author[1,2,3,5]{Dan Guo}[%
orcid=0000-0003-2594-254X,
email=guodan@hfut.edu.cn,
]
\cormark[1]

\address[1]{School of Computer Science and Information Engineering, School of Artificial Intelligence, Hefei University of Technology (HFUT)}
\address[2]{Key Laboratory of Knowledge Engineering with Big Data (HFUT), Ministry of Education}
\address[3]{Institute of Artificial Intelligence, Hefei Comprehensive National Science Center, China}
\address[4]{Zhejiang University, China}
\address[5]{Anhui Zhonghuitong Technology Co., Ltd.}

\cortext[1]{Corresponding author.}

\begin{abstract}
In this paper, we briefly introduce the solution developed by our team, HFUT-VUT, for the Micro-gesture Online Recognition track in the MiGA challenge at IJCAI 2024. The Micro-gesture Online Recognition task involves identifying the category and locating the start and end times of micro-gestures in video clips. Compared to the typical Temporal Action Detection task, the Micro-gesture Online Recognition task focuses more on distinguishing between micro-gestures and pinpointing the start and end times of actions. Our solution ranks \textbf{2nd} in the Micro-gesture Online Recognition track.  
\end{abstract}


\maketitle

\section{Introduction}

Humans can express emotions and communicate with others through various non-verbal forms, among which gestures play a crucial role in emotional expression and communication~\cite{chen2019analyze,chen2023smg,li2023joint,guo2024benchmarking,tang2022gloss}. Examples include ``cover face'', ``fold arms'', and ``cross fingers'', which convey human emotions to the outside world. Additionally, these micro-gestures (MGs) are often not spontaneous but occur unconsciously in specific environments. Unlike macro gestures intended for communication, non-spontaneous MGs better reflect genuine human emotions, making the study of MGs more meaningful in understanding human emotions. SMG~\cite{chen2023smg} and iMiGUE~\cite{liu2021imigue} are the datasets to assess and analyze human emotional states through MGs information. These datasets provide a stronger representation of human emotions, significantly contributing to a deeper understanding of genuine human feelings.

Compared to common macro gestures, Micro-gesture Online Recognition is more challenging because MGs appear more irregularly and randomly than existing action or gesture recognition datasets. Additionally, there may be co-occurrence relationships between different classes of actions, and transformations may occur between different 
MGs. Moreover, the finer distinctions between different categories of MGs make it more difficult to determine the start and end times of actions due to their smaller movement amplitudes.

In this challenge, we adopt PointTAD~\cite{tan2022pointtad} as the baseline. The main contributions of our method are as follows:
\begin{itemize}
\item We introduce the Mamba-MHSA block for Micro-gesture Online Recognition, which better distinguishes and locates action categories compared to the baseline model.
\item In the Micro-gesture Online Recognition challenge, our solution achieves an F1 score of 14.34 on the test set, securing 2nd in the competition. The experimental results demonstrate that our model can effectively distinguish and locate MGs.
\end{itemize}

\section{Related Work}

Current research predominantly focuses on common macro gestures or actions~\cite{li2021proposal,li2023vigt}, which have limited capability in reflecting human emotions. 
This is because humans can subjectively control their gestures and actions to hide their true emotions. In contrast, MGs typically occur involuntarily and uncontrollably, providing a more accurate reflection of genuine human emotions, which is crucial for understanding behavior and emotions. Here, we review the related technologies: micro-gesture datasets, temporal action detection, and Mamba.

\textbf{Micro-gesture Datasets.} The iMiGUE~\cite{liu2021imigue} dataset is the first publicly available dataset, aimed at recognizing and understanding suppressed or hidden emotions through MGs. 
It includes 359 videos with a total duration of 2092 minutes, collected from 72 subjects from 28 countries. The dataset is annotated with 18,499 MG samples across 32 categories, averaging 51 MG actions per video, with each MG instance ranging from 0.18 seconds to 80.92 seconds, and an average duration of 2.55 seconds. 
The SMG~\cite{chen2023smg} dataset focuses on naturally occurring MGs under stress, collected from 40 participants of various ages, genders, and racial backgrounds, divided into 16 types of MGs. The SMG dataset has been applied in various studies on micro-gesture recognition and emotion analysis, demonstrating its utility in these research fields. 

\textbf{Micro-gesture Online Recognition.} Guo \textit{et al.}~\cite{guo2023micro} proposed a novel deep network combining graph convolution and Transformer encoders to extract motion features from 2D skeleton sequences. This combination leverages the strengths of both graph convolution and Transformer. Their contributions collectively advance the state-of-the-art in micro-gesture recognition, providing a robust framework for emotion analysis based on MGs.

\textbf{Temporal Action Detection.} Temporal action detection has been studied as a multi-label frame-wise classification problem in previous literature. Early models~\cite{piergiovanni2018learning} mainly focused on modeling the temporal relationships between frames using Gaussian filters in the time dimension. Current research primarily deals with processing information at different scales and integrating spatiotemporal attention during processing. Tirupattur \textit{et al.}~\cite{tirupattur2021modeling} introduced an attention-based Multi-label Action Dependency layer (MLAD) in their model, significantly improving the co-occurrence dependencies and temporal dependencies of actions. 
Dai \textit{et al.}~\cite{dai2022ms} proposed a novel ConvtransFormer network named MS-TCT that incorporates global and local time relationship encoders and a time-scale mixer for effective multi-scale feature fusion~\cite{9446636}, addressing the complexities of temporal relationships. 
Tan \textit{et al.}~\cite{tan2022pointtad} presented an end-to-end action detection model named PointTAD that leverages learnable query points for precise localization and differentiation of actions in multi-label videos. These studies provide valuable insights for micro-gesture online recognition.

\textbf{Mamba.} The Transformer architecture and its core self-attention mechanism~\cite{wu2024waveformer,Wu_2023_CVPR,zhou2024advancing,wei2022robust} achieve significant success in deep learning. However, the Transformer faces inefficiency issues when processing long sequences. Structured State Space Models (SSMs)~\cite{gu2021efficiently}~\cite{gu2021combining}, combining characteristics of Recurrent Neural Networks (RNNs) and Convolutional Neural Networks (CNNs), have shown potential in certain data modalities. SSMs perform well on continuous signal data but less effectively on discrete and information-dense data. To address these shortcomings, Mamba introduces a selection mechanism that allows SSM parameters to adjust dynamically based on input data, improving model performance on discrete modalities. Mamba has notable advantages in inference speed and sequence length scalability. Thus, we incorporate Mamba into our model, combining Mamba~\cite{gu2023mamba}~\cite{shams2024ssamba} with self-attention to better model different semantics.

\section{Method}

\begin{figure*}[t!]
\centering
\includegraphics[width=1.0\linewidth]{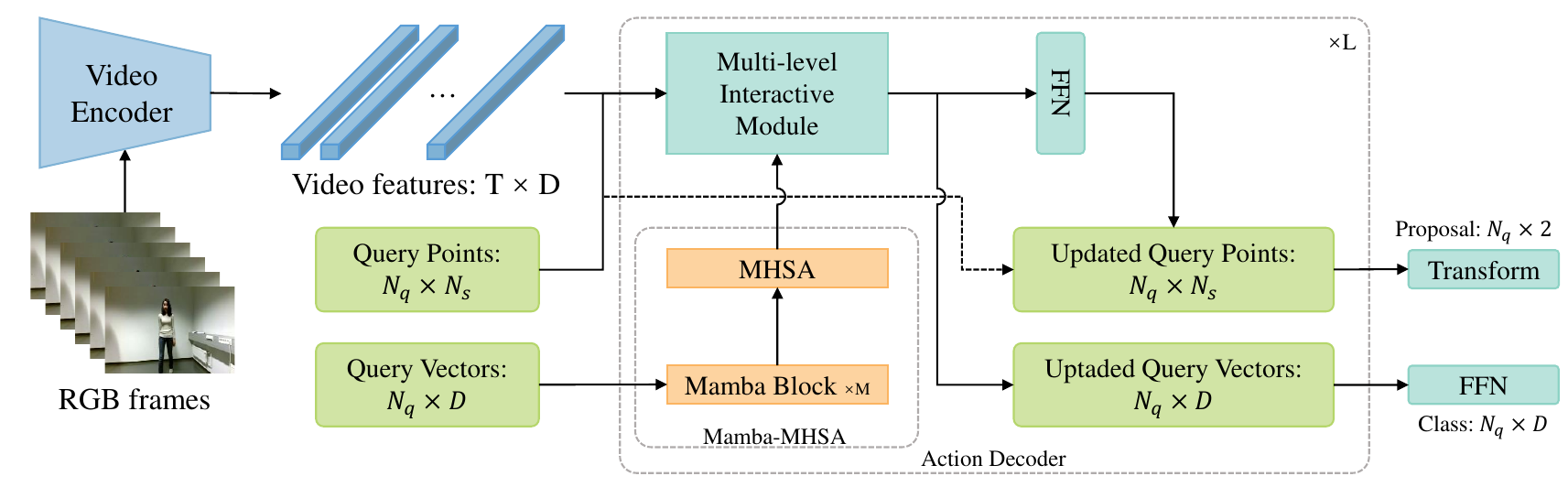}
\caption{The proposed model consists of a video encoder, which extracts video features from continuous RGB frames, and an action decoder.}
\label{fig:method}
\end{figure*}

\subsection{Task Definition}

We formulate the Micro-gesture Online Recognition task as a set prediction problem. Given a continuous video clip with $T$ frames, we predict a set of action instances $\phi = \left\{\phi_n = \left(t_n^s, t_n^e, c_n\right)\right\}_{n=1}^{N_q}$, where $N_q$ is the number of learnable queries, $t_n^s$ and $t_n^e$ are the starting and ending timestamps of the $n$-th detected instance, and $c_n$ is its action category. The ground truth action set to detect is denoted as $\hat{\phi} = \left\{\widehat{\phi_n} = \left(\widehat{t_n^s}, \widehat{t_n^e}, \widehat{c_n}\right)\right\}_{n=1}^{N_g}$, where $\widehat{t_n^s}$ and $\widehat{t_n^e}$ are the starting and ending timestamps of the $n$-th action, $\widehat{c_n}$ is the ground truth action category, and $N_g$ is the number of ground truth actions.

\subsection{Overall Architecture}

The overall architecture of our model is shown in Figure~\ref{fig:method}. The model consists of a video encoder and an action decoder. For each video sequence, we select an RGB sequence of length $T$, a set of learnable query points $P = \left\{P_i\right\}_{i=1}^{N_q}$, and query vectors $Q = \mathbb{R}^{T \times D}$. The learnable query points are used to locate the positions of action boundaries, and the query vectors decode action semantics and positions from the features input to the model. The action decoder comprises $L$ stacked decoder layers. Each layer of the action decoder takes video features, the latest query points $P$, and the latest query vectors $Q$ as input. Each action decoder layer includes two parts: 1) the Mamba-MHSA block models the relationships among query vectors and the potential relationships between different action categories; 2) the Multi-level Interactive Module dynamically models the relationships based on query vectors between point-level and same action categories. Finally, we use a Feed-Forward Network(FFN) to decode the action labels from the query vectors and convert the query points into detection outputs.

\subsection{Video Encoder}

We use the I3D network~\cite{carreira2017quo} as our model's video encoder, integrating the video encoder with the action decoder for end-to-end training. To facilitate model deployment and speed up feature extraction, we avoid using the optical flow part of the I3D backbone network. Finally, the temporal stride of the encoded video features is 4, and the spatiotemporal representations are compressed into temporal features through spatial average pooling.

\subsection{Learnable Query Points}

Using only the start and end times to represent an action instance limits its boundary and content description. Therefore, to improve the representation flexibility, a point-based representation method is used to learn keyframes of action boundaries and semantics within instances. 
For each query, the point-based representation is $P = \left\{t_i\right\}_{i=1}^{N_s}$, where $t_i$ is the time position of the $i$-th query point, and the number of points per query is $N_s$. During training, query points are initially placed at the midpoint of the input video sequence and are then refined through iterations in the action decoder layers by the query vectors $Q$, gradually approaching their final positions. 
Specifically, at each layer, the offsets of query points are predicted from the updated query vectors via linear projection. In action decoder layer $l$, the representation of a query's query points is $P^l = \left\{t_i^l\right\}_{i=1}^{N_s}$, with the offsets denoted as $\left\{\Delta t_i^l\right\}_{i=1}^{N_s}$. This operation can be summarized as:
\begin{equation}
P^{l+1} = \left\{\left(t_i^l + \Delta t_i^l \cdot s^l \cdot 0.5\right)\right\}_{i=1}^{N_s},
\label{eq:01}
\end{equation}
where $s^l = \max\left(t_i^l\right) - \min\left(t_i^l\right)$. For relatively short actions, the update step size of the query points is smaller, aiding in the localization of short actions. Additionally, the action query points updated by the previous action decoder layer become the input to the next action decoder layer after passing through a layer of FFN.

\subsection{Mamba-MHSA Block}
Compared to Transformers~\cite{vaswani2017attention,wang2024eulermormer,wang2024frequency}, the recently proposed Mamba has demonstrated powerful capabilities in sequence modeling. Therefore, we introduce Mamba into our model and combine it with the Multi-Head Self-Attention (MHSA) to model the relationships of query vectors, forming the Mamba-MHSA block. Our Mamba-MHSA module consists of $M$ of Mamba blocks and an MHSA. The Mamba block processes the query vectors $Q^m$ of the $m$-th Mamba block based on a selective state space model.

Mamba is designed based on state space models (SSMs) and requires defining three key parameters $A \in \mathbb{R}^{D \times D}$, $B \in \mathbb{R}^{D \times 1}$, and $C \in \mathbb{R}^{1 \times D}$. The SSMs are defined by the following differential equations:
\begin{equation}
 h'(t) = Ah(t) + BQ^m(t),
\label{eq:02}
\end{equation}
\begin{equation}
 y(t) = Ch(t).
\label{eq:03}
\end{equation}

We need to discretize the above equations. The discretized SSMs include a time parameter $\Delta$, which converts the continuous parameters $A$ and $B$ into discrete parameters. The specific formulas are as follows:
\begin{equation}
 A_x = \exp(\Delta A),
\label{eq:04}
\end{equation}
\begin{equation}
 B_x = (\Delta A)^{-1}(\exp(\Delta A) - I) \Delta A .
\label{eq:05}
\end{equation}

After discretization, the block can be expressed as:
\begin{equation}
 h_t = A_x h_{t-1} + B_x Q_t^m,
\label{eq:06}
\end{equation}
\begin{equation}
 y_t = Ch_t.
\label{eq:07}
\end{equation}

Next, we use a global convolution operation to obtain the output $Q^{m+1}$ by convolving the input sequence $Q^m$ with a structured convolutional kernel $K$. The convolution kernel $K$ is precomputed from the parameters $A$, $B$, and $C$, and its calculation method is as follows:
\begin{equation}
 Q^{m+1} = Mamba(Q^m) = Q^m \times K = Q^m \times (CB, CAB, \ldots, CA^{D-1}B).
\label{eq:08}
\end{equation}

After passing through $M$ of Mamba blocks, the query vectors $Q^M$ are input into a Multi-Head Self-Attention block to obtain the output. With the Mamba-MHSA block, the model gains stronger selectivity and perceptual capability for the input query vectors, allowing it to better model the relationships between different action instances.

\subsection{Multi-Level Interactive Module}

Previous temporal action detectors often have deficiencies in decoding sampled frames, as they typically aggregate semantics from different aspects and levels infrequently. Thus, we consider a multi-level interactive module to aggregate multi-level semantics.

\textbf{Point-Level Local Semantic Extraction} 
We use the deformable convolution~\cite{Wu_2023_CVPR1,9967838} to extract point-level features within a local neighborhood. For the $i$-th query point, considering that more time offsets can more precisely cover the area around the sub-points, thereby capturing more information, but they also increase the computational cost, we predict 4 time offsets $\left\{\Delta p_i\right\}_{i=1}^4$ and corresponding weights $\left\{w_i\right\}_{i=1}^4$ from the position of this point. Using the query point at frame $t_i$ as the center point, we add time offsets to form four deformable sub-points. These sub-points represent the local area around the center point. The features at the sub-points are extracted through bilinear interpolation and multiplied by the weight values to obtain the point-level feature $x_i$. This process can be represented as:
\begin{equation}
 x_i = \sum_{i=1}^{4}{\left(t_i + \Delta p_i\right) \times w_i}.
\label{eq:09}
\end{equation}

The offsets and weights are generated by linear projection from the query vector $q$. This process can be represented as:
\begin{equation}
\Delta q = \text{Linear}(q) \in \mathbb{R}^{N_q \times 4},
\label{eq:10}
\end{equation}
\begin{equation}
w = \text{Softmax}(\text{Linear}(q)) \in \mathbb{R}^{N_q \times 4}.
\label{eq:11}
\end{equation}

\textbf{Instance-Level Semantic Mixing} Since actions can occur simultaneously, modeling only the temporal aspect may cause overlapping actions to have similar representations, leading to classification errors. Therefore, dynamic convolution is used to mix semantics across frames and channels. The mixed features of the query points use $x \in \mathbb{R}^{N_s \times D}$. Given the query vector $q$, the parameters for frame mix and channel mix are generated:
\begin{equation}
 \theta_f = \text{Linear}(q) \in \mathbb{R}^{N_s \times N_s} ,
 \theta_{c,1} = \text{Linear}(q) \in \mathbb{R}^{D \times D'} ,
 \theta_{c,2} = \text{Linear}(q) \in \mathbb{R}^{D' \times D}. 
\label{eq:12}
\end{equation}

Frame mix is performed by projecting and then activating with LayerNorm and ReLU across $N_s$ points to explore intra-instance relationships:
\begin{equation}
 x_f = \text{ReLU}(\text{LayerNorm}(x^T \theta_f)) \in \mathbb{R}^{D \times N_s}.
\label{eq:13}
\end{equation}

Channel mix enhances action semantics using dynamic projection along the channel dimension:
\begin{equation}
 x_c = \text{ReLU}(\text{LayerNorm}(\text{ReLU}(\text{LayerNorm}(x \theta_{c,1})) \theta_{c,2})) \in \mathbb{R}^{N_s \times D}.
\label{eq:14}
\end{equation}

These two features are then concatenated along the channel and compressed through a linear layer to the size of the query vector. The query vector is updated to obtain the query vector for the next layer input $q^{l+1}$. This process can be represented as:

\begin{equation}
 q^{l+1} = q^l + \text{Linear}(\text{Concat}(x_f^T, x_c)).
\label{eq:14}
\end{equation}

\section{Experiments}
\subsection{Dataset and Evaluation Metric}

\textbf{Dataset.} The spontaneous Micro-Gesture (SMG) dataset~\cite{chen2023smg} consists of 3,692 samples of 17 MGs. The dataset employs a cross-subject evaluation protocol by dividing the 40 subjects into a training group consisting of long sequences from 35 subjects and a testing group of sequences from 5 subjects. We only use RGB sequences as input.

\noindent\textbf{Evaluation Metric.} We jointly evaluate the detection and classification performances of algorithms using the $F1$ score measurement defined below:
\begin{equation}
 F1 = 2 \cdot \frac{ \text{Precision} \cdot \text{Recall}}{\text{Precision} + \text{Recall}}.
\label{eq:15}
\end{equation}

Given a long video sequence that needs to be evaluated, Precision is the fraction of correctly classified MGs among all gestures retrieved in the sequence by the algorithms, while Recall (or sensitivity) is the fraction of MGs that have been correctly retrieved over the total amount of annotated MGs.

\subsection{Implementation Details}

We use the I3D backbone network to extract video frames at a rate of 10 fps. A sliding window mechanism is employed to preprocess video sequences, with the window size($\beta$) set to 128 frames to accommodate most action categories. During training, the overlap ratio is set to 0.75, while for inference, the overlap ratio is 0. We set $N_q$ to 48 and $N_s$ to 30. The I3D backbone uses pre-trained weights from Kinetics400~\cite{kay2017kinetics}. The batch size is set to 1, and the initial learning rate is 1e-4, halved every 10 epochs, for a total of 50 epochs. 

\begin{table}[htbp]
\centering
\caption{The top-3 results of Micro-gesture Online Recognition on the SMG test set. Data is provided by the Kaggle competition page\protect\footnotemark[1].
}
\begin{tabular}{c|cc}
\toprule
Rank  & Team  & F1 Score \\
\midrule
1     & NPU-MUCIS & 27.57 \\
\midrule
2     & \textbf{HFUT-VUT(ours)} & \textbf{14.34} \\
\midrule
3     & JDY203 & 9.28 \\
\bottomrule
\end{tabular}
\label{tab:Table1}
\end{table}
\footnotetext[1]{The Kaggle competition page: \href{https://www.kaggle.com/competitions/2nd-miga-ijcai-challenge-track2/leaderboard}{https://www.kaggle.com/competitions/2nd-miga-ijcai-challenge-track2/leaderboard}}

\subsection{Experimental Results}
As shown in Table~\ref{tab:Table1}, we report the results of the top three teams on the SMG dataset test set. Our team secured the second place. Although there remains a notable performance disparity between our method and the first-place ``NPU-MUCIS'' team, our method significantly exceeds the performance of the third-place ``JDY203'' team by 54.52\%.

\subsection{Ablation Study}

Study on the Number of Query Points ($\alpha$). 
In Table~\ref{tab:Table2a}, we conduct an ablation study on different numbers of query points. We observe that the model's performance improves as the number of query points increases when the number is less than 30. However, when the number of query points exceeds 30, the model's performance starts to decrease. Therefore, we choose 30 as the default number of query points for our model.

Study on Window Size ($\beta$). We examine the potential impact of different window sizes on the model results. We consider five different initializations for window size to accommodate the majority of action lengths in the dataset. 
As shown in Table~\ref{tab:Table2b}, the model achieves the best performance when the window size is set to 128. Thus, we set the window size to 128.

Study on the number of layers in the Action Decoder ($L$). We investigate the influence of different numbers of layers in the action decoder on the model. According to the results in Table~\ref{tab:Table2c}, increasing the number of layers in the action decoder allows the model to learn deeper information, thereby improving its performance. However, when the number of layers exceeds 4, the model's performance begins to decrease.

\begin{table}[t!]
\centering
\caption{The ablation experiments of our method on the SMG dataset.}
\captionsetup[subtable]{labelformat=simple,labelsep=period}
\begin{subtable}[t]{0.24\linewidth}
\centering
\caption{Query Points in action detectors parameter $\alpha$}
\begin{tabular}{cccc}
\toprule
$\alpha$ & F1-score \\
\midrule
25    & 11.33 \\
27    & 13.48 \\
\textbf{30} & \textbf{14.34} \\
31    & 13.49 \\
32    & 13.43 \\
35    & 9.81 \\
\bottomrule
\end{tabular}%
\label{tab:Table2a}%
\end{subtable}
\hfill
\begin{subtable}[t]{0.24\linewidth}
\centering
\caption{Window size in action detectors parameter $\beta$}
\begin{tabular}{cccc}
\toprule
$\beta$ & F1-score \\
\midrule
16    & 7.33 \\
32    & 8.45 \\
64    & 9.27 \\
\textbf{128} & \textbf{14.34} \\
200   & 7.95 \\
\bottomrule
\end{tabular}%
\label{tab:Table2b}%
\end{subtable}
\hfill
\begin{subtable}[t]{0.24\linewidth}
\centering
\caption{Action decoder parameter $L$}
\begin{tabular}{cccc}
\toprule
$L$     & F1-score \\
\midrule
2     & 8.3 \\
3     & 11.96 \\
\textbf{4} & \textbf{14.34} \\
5     & 10.01 \\
\bottomrule
\end{tabular}%
\label{tab:Table2c}%
\end{subtable}
\hfill
\begin{subtable}[t]{0.24\linewidth}
\centering
\caption{Mamba Block parameter $M$}
\begin{tabular}{cccc}
\toprule
$M$     & F1-score \\
\midrule
1     & 8.42 \\
\textbf{2} & \textbf{14.34} \\
3     & 9.22 \\
\bottomrule
\end{tabular}%
\label{tab:Table2d}%
\end{subtable}
\end{table}

Study on the number of Mamba Blocks ($M$). To balance computational resources, we study the impact of the number of Mamba blocks on the model. As indicated in Table~\ref{tab:Table2d}, the model performs best when $M$ is set to 2. Additionally, when the number of Mamba blocks exceeds 2, the model encounters issues with gradient explosion.

\section{Conclusion}
In this paper, we present a solution for the Micro-gesture Online Recognition (MiGA) challenge at IJCAI 2024. Our approach is based on the PointTAD baseline, enhanced with Mamba-MHSA to improve the model's ability to model sequences. This module effectively enhances the model's capability for Micro-gesture Online Recognition, achieving an experimental result of 14.34 on the SMG dataset. 
In future work, we will consider incorporating skeletal data into the model to enhance its recognition ability for Micro-gesture Online Recognition.

\begin{acknowledgments}
This work was supported by the National Key R\&D Program of China (NO.2022YFB4500601), the National Natural Science Foundation of China (62272144,72188101,62020106007 and U20A20183), the Major Project of Anhui Province (202203a05020011), and the Fundamental Research Funds for the Central Universities. 
\end{acknowledgments}

\bibliography{sample-ceur}

\begin{thebibliography}{29}
\expandafter\ifx\csname natexlab\endcsname\relax\def\natexlab#1{#1}\fi
\providecommand{\url}[1]{\texttt{#1}}
\providecommand{\href}[2]{#2}
\providecommand{\path}[1]{#1}
\providecommand{\DOIprefix}{doi:}
\providecommand{\ArXivprefix}{arXiv:}
\providecommand{\URLprefix}{URL: }
\providecommand{\Pubmedprefix}{pmid:}
\providecommand{\doi}[1]{\href{http://dx.doi.org/#1}{\path{#1}}}
\providecommand{\Pubmed}[1]{\href{pmid:#1}{\path{#1}}}
\providecommand{\bibinfo}[2]{#2}
\ifx\xfnm\relax \def\xfnm[#1]{\unskip,\space#1}\fi
\bibitem[{Chen et~al.(2019)Chen, Liu, Li, Shi, and Zhao}]{chen2019analyze}
\bibinfo{author}{H.~Chen}, \bibinfo{author}{X.~Liu}, \bibinfo{author}{X.~Li},
  \bibinfo{author}{H.~Shi}, \bibinfo{author}{G.~Zhao},
\newblock \bibinfo{title}{Analyze spontaneous gestures for emotional stress
  state recognition: A micro-gesture dataset and analysis with deep learning},
\newblock in: \bibinfo{booktitle}{2019 14th IEEE International Conference on
  Automatic Face \& Gesture Recognition (FG 2019)}, \bibinfo{year}{2019}, pp.
  \bibinfo{pages}{1--8}.
\bibitem[{Chen et~al.(2023)Chen, Shi, Liu, Li, and Zhao}]{chen2023smg}
\bibinfo{author}{H.~Chen}, \bibinfo{author}{H.~Shi}, \bibinfo{author}{X.~Liu},
  \bibinfo{author}{X.~Li}, \bibinfo{author}{G.~Zhao},
\newblock \bibinfo{title}{Smg: A micro-gesture dataset towards spontaneous body
  gestures for emotional stress state analysis},
\newblock \bibinfo{journal}{International Journal of Computer Vision}
  \bibinfo{volume}{131} (\bibinfo{year}{2023}) \bibinfo{pages}{1346--1366}.
\bibitem[{Li et~al.(2023)Li, Guo, Chen, Peng, and Wang}]{li2023joint}
\bibinfo{author}{K.~Li}, \bibinfo{author}{D.~Guo}, \bibinfo{author}{G.~Chen},
  \bibinfo{author}{X.~Peng}, \bibinfo{author}{M.~Wang},
\newblock \bibinfo{title}{Joint skeletal and semantic embedding loss for
  micro-gesture classification},
\newblock \bibinfo{journal}{arXiv preprint arXiv:2307.10624}
  (\bibinfo{year}{2023}).
\bibitem[{Guo et~al.(2024)Guo, Li, Hu, Zhang, and Wang}]{guo2024benchmarking}
\bibinfo{author}{D.~Guo}, \bibinfo{author}{K.~Li}, \bibinfo{author}{B.~Hu},
  \bibinfo{author}{Y.~Zhang}, \bibinfo{author}{M.~Wang},
\newblock \bibinfo{title}{Benchmarking micro-action recognition: Dataset,
  methods, and applications},
\newblock \bibinfo{journal}{IEEE Transactions on Circuits and Systems for Video
  Technology}  (\bibinfo{year}{2024}).
\bibitem[{Tang et~al.(2022)Tang, Hong, Guo, and Wang}]{tang2022gloss}
\bibinfo{author}{S.~Tang}, \bibinfo{author}{R.~Hong}, \bibinfo{author}{D.~Guo},
  \bibinfo{author}{M.~Wang},
\newblock \bibinfo{title}{Gloss semantic-enhanced network with online
  back-translation for sign language production},
\newblock in: \bibinfo{booktitle}{Proceedings of the 30th ACM International
  Conference on Multimedia}, \bibinfo{year}{2022}, pp.
  \bibinfo{pages}{5630--5638}.
\bibitem[{Liu et~al.(2021)Liu, Shi, Chen, Yu, Li, and Zhao}]{liu2021imigue}
\bibinfo{author}{X.~Liu}, \bibinfo{author}{H.~Shi}, \bibinfo{author}{H.~Chen},
  \bibinfo{author}{Z.~Yu}, \bibinfo{author}{X.~Li}, \bibinfo{author}{G.~Zhao},
\newblock \bibinfo{title}{imigue: An identity-free video dataset for
  micro-gesture understanding and emotion analysis},
\newblock in: \bibinfo{booktitle}{Proceedings of the IEEE/CVF Conference on
  Computer Vision and Pattern Recognition}, \bibinfo{year}{2021}, pp.
  \bibinfo{pages}{10631--10642}.
\bibitem[{Tan et~al.(2022)Tan, Zhao, Shi, Kang, and Wang}]{tan2022pointtad}
\bibinfo{author}{J.~Tan}, \bibinfo{author}{X.~Zhao}, \bibinfo{author}{X.~Shi},
  \bibinfo{author}{B.~Kang}, \bibinfo{author}{L.~Wang},
\newblock \bibinfo{title}{Pointtad: Multi-label temporal action detection with
  learnable query points},
\newblock \bibinfo{journal}{Advances in Neural Information Processing Systems}
  \bibinfo{volume}{35} (\bibinfo{year}{2022}) \bibinfo{pages}{15268--15280}.
\bibitem[{Li et~al.(2021)Li, Guo, and Wang}]{li2021proposal}
\bibinfo{author}{K.~Li}, \bibinfo{author}{D.~Guo}, \bibinfo{author}{M.~Wang},
\newblock \bibinfo{title}{Proposal-free video grounding with contextual pyramid
  network},
\newblock in: \bibinfo{booktitle}{Proceedings of the AAAI Conference on
  Artificial Intelligence}, \bibinfo{year}{2021}, pp.
  \bibinfo{pages}{1902--1910}.
\bibitem[{Li et~al.(2023)Li, Guo, and Wang}]{li2023vigt}
\bibinfo{author}{K.~Li}, \bibinfo{author}{D.~Guo}, \bibinfo{author}{M.~Wang},
\newblock \bibinfo{title}{Vigt: proposal-free video grounding with a learnable
  token in the transformer},
\newblock \bibinfo{journal}{Science China Information Sciences}
  \bibinfo{volume}{66} (\bibinfo{year}{2023}) \bibinfo{pages}{202102}.
\bibitem[{Guo et~al.(2023)Guo, Peng, Huang, and Xia}]{guo2023micro}
\bibinfo{author}{X.~Guo}, \bibinfo{author}{W.~Peng},
  \bibinfo{author}{H.~Huang}, \bibinfo{author}{Z.~Xia},
\newblock \bibinfo{title}{Micro-gesture online recognition with
  graph-convolution and multiscale transformers for long sequence}
  (\bibinfo{year}{2023}).
\bibitem[{Piergiovanni and Ryoo(2018)}]{piergiovanni2018learning}
\bibinfo{author}{A.~Piergiovanni}, \bibinfo{author}{M.~S. Ryoo},
\newblock \bibinfo{title}{Learning latent super-events to detect multiple
  activities in videos},
\newblock in: \bibinfo{booktitle}{Proceedings of the IEEE Conference on
  Computer Vision and Pattern Recognition}, \bibinfo{year}{2018}, pp.
  \bibinfo{pages}{5304--5313}.
\bibitem[{Tirupattur et~al.(2021)Tirupattur, Duarte, Rawat, and
  Shah}]{tirupattur2021modeling}
\bibinfo{author}{P.~Tirupattur}, \bibinfo{author}{K.~Duarte},
  \bibinfo{author}{Y.~S. Rawat}, \bibinfo{author}{M.~Shah},
\newblock \bibinfo{title}{Modeling multi-label action dependencies for temporal
  action localization},
\newblock in: \bibinfo{booktitle}{Proceedings of the IEEE/CVF Conference on
  Computer Vision and Pattern Recognition}, \bibinfo{year}{2021}, pp.
  \bibinfo{pages}{1460--1470}.
\bibitem[{Dai et~al.(2022)Dai, Das, Kahatapitiya, Ryoo, and
  Br{\'e}mond}]{dai2022ms}
\bibinfo{author}{R.~Dai}, \bibinfo{author}{S.~Das},
  \bibinfo{author}{K.~Kahatapitiya}, \bibinfo{author}{M.~S. Ryoo},
  \bibinfo{author}{F.~Br{\'e}mond},
\newblock \bibinfo{title}{Ms-tct: Multi-scale temporal convtransformer for
  action detection},
\newblock in: \bibinfo{booktitle}{Proceedings of the IEEE/CVF Conference on
  Computer Vision and Pattern Recognition}, \bibinfo{year}{2022}, pp.
  \bibinfo{pages}{20041--20051}.
\bibitem[{Wu et~al.(2021)Wu, Zhang, Xuan, Yang, and Yan}]{9446636}
\bibinfo{author}{Z.~Wu}, \bibinfo{author}{K.~Zhang}, \bibinfo{author}{H.~Xuan},
  \bibinfo{author}{J.~Yang}, \bibinfo{author}{Y.~Yan},
\newblock \bibinfo{title}{Dapc-net: Deformable alignment and pyramid context
  completion networks for video inpainting},
\newblock \bibinfo{journal}{IEEE Signal Processing Letters}
  \bibinfo{volume}{28} (\bibinfo{year}{2021}) \bibinfo{pages}{1145--1149}.
\bibitem[{Wu et~al.(2024)Wu, Sun, Xuan, Liu, and Yan}]{wu2024waveformer}
\bibinfo{author}{Z.~Wu}, \bibinfo{author}{C.~Sun}, \bibinfo{author}{H.~Xuan},
  \bibinfo{author}{G.~Liu}, \bibinfo{author}{Y.~Yan},
\newblock \bibinfo{title}{Waveformer: Wavelet transformer for noise-robust
  video inpainting},
\newblock in: \bibinfo{booktitle}{Proceedings of the AAAI Conference on
  Artificial Intelligence}, volume~\bibinfo{volume}{38}, \bibinfo{year}{2024},
  pp. \bibinfo{pages}{6180--6188}.
\bibitem[{Wu et~al.(2023)Wu, Sun, Xuan, and Yan}]{Wu_2023_CVPR}
\bibinfo{author}{Z.~Wu}, \bibinfo{author}{C.~Sun}, \bibinfo{author}{H.~Xuan},
  \bibinfo{author}{Y.~Yan},
\newblock \bibinfo{title}{Deep stereo video inpainting},
\newblock in: \bibinfo{booktitle}{Proceedings of the IEEE/CVF Conference on
  Computer Vision and Pattern Recognition}, \bibinfo{year}{2023}, pp.
  \bibinfo{pages}{5693--5702}.
\bibitem[{Zhou et~al.(2024)Zhou, Guo, Zhong, and Wang}]{zhou2024advancing}
\bibinfo{author}{J.~Zhou}, \bibinfo{author}{D.~Guo},
  \bibinfo{author}{Y.~Zhong}, \bibinfo{author}{M.~Wang},
\newblock \bibinfo{title}{Advancing weakly-supervised audio-visual video
  parsing via segment-wise pseudo labeling},
\newblock \bibinfo{journal}{arXiv preprint arXiv:2406.00919}
  (\bibinfo{year}{2024}).
\bibitem[{Wei et~al.(2022)Wei, Zhang, Xu, Hong, Fan, and Yan}]{wei2022robust}
\bibinfo{author}{Y.~Wei}, \bibinfo{author}{Z.~Zhang}, \bibinfo{author}{M.~Xu},
  \bibinfo{author}{R.~Hong}, \bibinfo{author}{J.~Fan},
  \bibinfo{author}{S.~Yan},
\newblock \bibinfo{title}{Robust attention deraining network for synchronous
  rain streaks and raindrops removal},
\newblock in: \bibinfo{booktitle}{Proceedings of the 30th ACM International
  Conference on Multimedia}, \bibinfo{year}{2022}, pp.
  \bibinfo{pages}{6464--6472}.
\bibitem[{Gu et~al.(2021{\natexlab{a}})Gu, Goel, and
  R{\'e}}]{gu2021efficiently}
\bibinfo{author}{A.~Gu}, \bibinfo{author}{K.~Goel},
  \bibinfo{author}{C.~R{\'e}},
\newblock \bibinfo{title}{Efficiently modeling long sequences with structured
  state spaces},
\newblock \bibinfo{journal}{arXiv preprint arXiv:2111.00396}
  (\bibinfo{year}{2021}{\natexlab{a}}).
\bibitem[{Gu et~al.(2021{\natexlab{b}})Gu, Johnson, Goel, Saab, Dao, Rudra, and
  R{\'e}}]{gu2021combining}
\bibinfo{author}{A.~Gu}, \bibinfo{author}{I.~Johnson},
  \bibinfo{author}{K.~Goel}, \bibinfo{author}{K.~Saab},
  \bibinfo{author}{T.~Dao}, \bibinfo{author}{A.~Rudra},
  \bibinfo{author}{C.~R{\'e}},
\newblock \bibinfo{title}{Combining recurrent, convolutional, and
  continuous-time models with linear state space layers},
\newblock \bibinfo{journal}{Advances in neural information processing systems}
  \bibinfo{volume}{34} (\bibinfo{year}{2021}{\natexlab{b}})
  \bibinfo{pages}{572--585}.
\bibitem[{Gu and Dao(2023)}]{gu2023mamba}
\bibinfo{author}{A.~Gu}, \bibinfo{author}{T.~Dao},
\newblock \bibinfo{title}{Mamba: Linear-time sequence modeling with selective
  state spaces},
\newblock \bibinfo{journal}{arXiv preprint arXiv:2312.00752}
  (\bibinfo{year}{2023}).
\bibitem[{Shams et~al.(2024)Shams, Dindar, Jiang, and
  Mesgarani}]{shams2024ssamba}
\bibinfo{author}{S.~Shams}, \bibinfo{author}{S.~S. Dindar},
  \bibinfo{author}{X.~Jiang}, \bibinfo{author}{N.~Mesgarani},
\newblock \bibinfo{title}{Ssamba: Self-supervised audio representation learning
  with mamba state space model},
\newblock \bibinfo{journal}{arXiv preprint arXiv:2405.11831}
  (\bibinfo{year}{2024}).
\bibitem[{Carreira and Zisserman(2017)}]{carreira2017quo}
\bibinfo{author}{J.~Carreira}, \bibinfo{author}{A.~Zisserman},
\newblock \bibinfo{title}{Quo vadis, action recognition? a new model and the
  kinetics dataset},
\newblock in: \bibinfo{booktitle}{proceedings of the IEEE Conference on
  Computer Vision and Pattern Recognition}, \bibinfo{year}{2017}, pp.
  \bibinfo{pages}{6299--6308}.
\bibitem[{Vaswani et~al.(2017)Vaswani, Shazeer, Parmar, Uszkoreit, Jones,
  Gomez, Kaiser, and Polosukhin}]{vaswani2017attention}
\bibinfo{author}{A.~Vaswani}, \bibinfo{author}{N.~Shazeer},
  \bibinfo{author}{N.~Parmar}, \bibinfo{author}{J.~Uszkoreit},
  \bibinfo{author}{L.~Jones}, \bibinfo{author}{A.~N. Gomez},
  \bibinfo{author}{{\L}.~Kaiser}, \bibinfo{author}{I.~Polosukhin},
\newblock \bibinfo{title}{Attention is all you need},
\newblock \bibinfo{journal}{Proceedings of the Advances in Neural Information
  Processing Systems} \bibinfo{volume}{30} (\bibinfo{year}{2017}).
\bibitem[{Wang et~al.(2024{\natexlab{a}})Wang, Guo, Li, and
  Wang}]{wang2024eulermormer}
\bibinfo{author}{F.~Wang}, \bibinfo{author}{D.~Guo}, \bibinfo{author}{K.~Li},
  \bibinfo{author}{M.~Wang},
\newblock \bibinfo{title}{Eulermormer: Robust eulerian motion magnification via
  dynamic filtering within transformer},
\newblock in: \bibinfo{booktitle}{Proceedings of the AAAI Conference on
  Artificial Intelligence}, volume~\bibinfo{volume}{38},
  \bibinfo{year}{2024}{\natexlab{a}}, pp. \bibinfo{pages}{5345--5353}.
\bibitem[{Wang et~al.(2024{\natexlab{b}})Wang, Guo, Li, Zhong, and
  Wang}]{wang2024frequency}
\bibinfo{author}{F.~Wang}, \bibinfo{author}{D.~Guo}, \bibinfo{author}{K.~Li},
  \bibinfo{author}{Z.~Zhong}, \bibinfo{author}{M.~Wang},
\newblock \bibinfo{title}{Frequency decoupling for motion magnification via
  multi-level isomorphic architecture},
\newblock \bibinfo{journal}{arXiv preprint arXiv:2403.07347}
  (\bibinfo{year}{2024}{\natexlab{b}}).
\bibitem[{Wu et~al.(2023{\natexlab{a}})Wu, Xuan, Sun, Guan, Zhang, and
  Yan}]{Wu_2023_CVPR1}
\bibinfo{author}{Z.~Wu}, \bibinfo{author}{H.~Xuan}, \bibinfo{author}{C.~Sun},
  \bibinfo{author}{W.~Guan}, \bibinfo{author}{K.~Zhang},
  \bibinfo{author}{Y.~Yan},
\newblock \bibinfo{title}{Semi-supervised video inpainting with cycle
  consistency constraints},
\newblock in: \bibinfo{booktitle}{Proceedings of the IEEE/CVF Conference on
  Computer Vision and Pattern Recognition (CVPR)},
  \bibinfo{year}{2023}{\natexlab{a}}, pp. \bibinfo{pages}{22586--22595}.
\bibitem[{Wu et~al.(2023{\natexlab{b}})Wu, Sun, Xuan, Zhang, and Yan}]{9967838}
\bibinfo{author}{Z.~Wu}, \bibinfo{author}{C.~Sun}, \bibinfo{author}{H.~Xuan},
  \bibinfo{author}{K.~Zhang}, \bibinfo{author}{Y.~Yan},
\newblock \bibinfo{title}{Divide-and-conquer completion network for video
  inpainting},
\newblock \bibinfo{journal}{IEEE Transactions on Circuits and Systems for Video
  Technology} \bibinfo{volume}{33} (\bibinfo{year}{2023}{\natexlab{b}})
  \bibinfo{pages}{2753--2766}.
\bibitem[{Kay et~al.(2017)Kay, Carreira, Simonyan, Zhang, Hillier,
  Vijayanarasimhan, Viola, Green, Back, Natsev et~al.}]{kay2017kinetics}
\bibinfo{author}{W.~Kay}, \bibinfo{author}{J.~Carreira},
  \bibinfo{author}{K.~Simonyan}, \bibinfo{author}{B.~Zhang},
  \bibinfo{author}{C.~Hillier}, \bibinfo{author}{S.~Vijayanarasimhan},
  \bibinfo{author}{F.~Viola}, \bibinfo{author}{T.~Green},
  \bibinfo{author}{T.~Back}, \bibinfo{author}{P.~Natsev}, et~al.,
\newblock \bibinfo{title}{The kinetics human action video dataset},
\newblock \bibinfo{journal}{arXiv preprint arXiv:1705.06950}
  (\bibinfo{year}{2017}).

\end{thebibliography}

\end{document}